\documentclass[sigconf,nonacm]{acmart}

\setcopyright{none}
\renewcommand\footnotetextcopyrightpermission[1]{}
\makeatletter
\def\@ACM@checkaffil{}
\makeatother

\usepackage{booktabs}
\usepackage{enumitem}
\usepackage{tikz}
\usetikzlibrary{arrows.meta,positioning,calc,matrix}
\setlist[enumerate]{leftmargin=1.15em,itemsep=0.4pt,topsep=1.0pt,parsep=0pt,partopsep=0pt}
\setlist[itemize]{label=$\bullet$,leftmargin=1.05em,itemsep=0.4pt,topsep=1.0pt,parsep=0pt,partopsep=0pt}
\renewcommand{\textbullet}{\ensuremath{\bullet}}
\begin{document}

\title{Provenance Loss and Repair in Multilingual Compaction}

\author{Ian Miao}
\affiliation{%
  \institution{Google}
}
\email{yilinmiao@google.com}

\author{Fangrui Zhu}
\affiliation{%
  \institution{Google}
}
\email{fangrui@google.com}

\author{Bowen Jiang}
\affiliation{%
  \institution{University of Pennsylvania}
}
\email{bwjiang@alumni.upenn.edu}

\begin{abstract}
Conversational assistants compact older messages into short notes and evict the originals. When the note's language requires a distinction that English leaves open, such as Spanish adjective gender (\texttt{agotado} vs.\ \texttt{agotada}), messages that omit or explicitly state that attribute can yield the same note. After eviction, later turns cannot tell them apart: taking the inflected form literally asserts an attribute the user never gave, whereas discounting it as routine grammar discards one the user did give. We propose a training-free repair across writing, pre-eviction verification and reading. The writer marks unstated attributes with paired forms (e.g., \texttt{agotado/a}), a likelihood check before eviction rewrites unsupported single forms into paired ones, and the reader trusts remaining single forms as stated facts. On five \texttt{Qwen3.5} and \texttt{Gemma-4} models, eviction cuts recall of stated attributes from $90\%$ to $34\%$ in question answering (QA); our repair raises F1 (which penalizes both errors) on every model, by $9$--$53$ points in QA and $12$--$40$ points in agent handovers.
\end{abstract}


\maketitle

\section{Introduction}
\label{sec:intro}
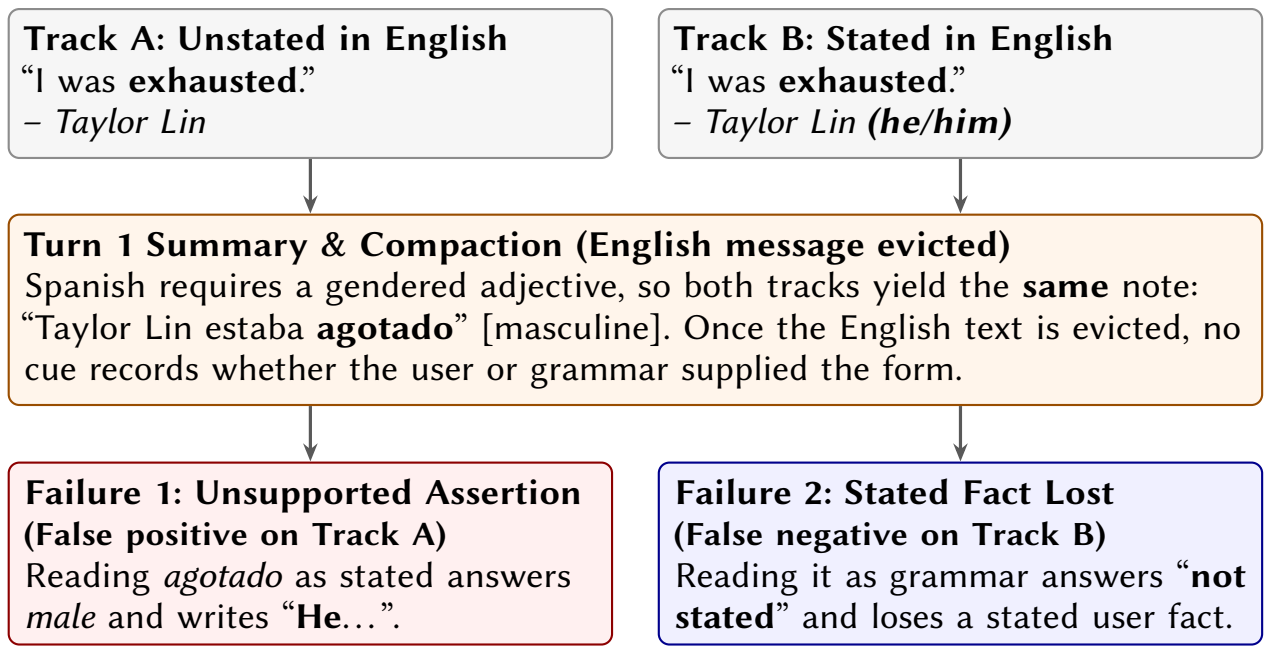
\begin{figure}[t]
\centering
\begin{tikzpicture}[
  font=\fontsize{6.9}{7.8}\selectfont\sffamily,
  box/.style={draw=black!45, rounded corners=2pt, inner sep=3.0pt, align=left},
  trka/.style={box, fill=gray!7, text width=3.78cm},
  trkb/.style={box, fill=gray!7, text width=3.78cm},
  mid/.style={box, fill=orange!8, draw=orange!60!black, text width=8.08cm},
  err1/.style={box, fill=red!6, draw=red!55!black, text width=3.78cm},
  err2/.style={box, fill=blue!6, draw=blue!55!black, text width=3.78cm},
  arr/.style={-{Stealth[length=3.4pt,width=2.7pt]}, semithick, draw=black!65}
]

\node[trka, anchor=north west] (a) at (0,0) {
  \textbf{Track A: Unstated in English}\\
  ``I was \textbf{exhausted}.''\\
  \emph{-- Taylor Lin}
};

\node[trkb, anchor=north west] (b) at (4.30cm,0) {
  \textbf{Track B: Stated in English}\\
  ``I was \textbf{exhausted}.''\\
  \emph{-- Taylor Lin \textbf{(he/him)}}
};

\node[mid, anchor=north west] (m) at (0,-1.36cm) {
  \textbf{Turn 1 Summary \& Compaction (English message evicted)}\\
  Spanish requires a gendered adjective, so both tracks yield the \textbf{same} note: ``Taylor Lin estaba \textbf{agotado}'' {[masculine]}. Once the English text is evicted, no cue records whether the user or grammar supplied the form.
};

\node[err1, anchor=north west] (e1) at (0,-3.00cm) {
  \textbf{Failure 1: Unsupported Assertion}\\
  {\fontsize{6.4}{7.1}\selectfont\textbf{(False positive on Track A)}}\\
  Reading \emph{agotado} as stated answers \emph{male} and writes ``\textbf{He}\dots''.
};

\node[err2, anchor=north west] (e2) at (4.30cm,-3.00cm) {
  \textbf{Failure 2: Stated Fact Lost}\\
  {\fontsize{6.4}{7.1}\selectfont\textbf{(False negative on Track B)}}\\
  Reading it as grammar answers ``\textbf{not stated}'' and loses a stated user fact.
};

\draw[arr] (a.south) -- (a.south |- m.north);
\draw[arr] (b.south) -- (b.south |- m.north);
\draw[arr] (e1.north |- m.south) -- (e1.north);
\draw[arr] (e2.north |- m.south) -- (e2.north);

\end{tikzpicture}
\Description{Single-column diagram contrasting Track A (unstated English message: I was exhausted) and Track B (the same message from a sender tagged he/him). Both tracks produce the same Spanish note with the masculine form agotado once the English message is evicted, creating coupled provenance failures at Turn 2: taking the form as user-stated causes Failure 1 (unsupported assertion on Track A), while discounting it causes Failure 2 (stated fact lost on Track B).}
\caption{Coupled provenance failures after compaction: unstated (Track~A) and stated (Track~B) inputs yield the same inflected note once evicted, so any fixed reader either asserts unstated attributes or loses stated ones.}
\label{fig:overview}
\end{figure}

Conversational agents bound context length by compressing earlier context \citep{jiang2023llmlingua,xiao2023streamingllm,li2024snapkv} or summarizing turns into notes kept across sessions \citep{packer2023memgpt,chhikara2025mem0}. This compaction often crosses languages: in customer support, an English intake message is distilled into a local-language note for downstream agents, local tools and regional staff; once the raw message is evicted, later turns read only that note. Yet many target languages require distinctions that English leaves open: Romance adjectives inflect for gender (\emph{agotado}/\emph{agotada} in Spanish) \citep{corbett1991gender}, while Mandarin kinship nouns split maternal vs.\ paternal lineage (\emph{waipo}/\emph{nainai} for grandmother) and relative age (\emph{gege}/\emph{didi} for older vs.\ younger brother) \citep{li2024kinship}. For such binary contrasts, a note committing to one option (\emph{agotado}, \emph{waipo}) is \emph{single-form}, whereas one keeping both (\texttt{agotado/a}, \emph{waipo/nainai}) is \emph{paired} \citep{sczesny2016can}.

Eviction creates two coupled provenance failures (Fig.~\ref{fig:overview}). When the attribute is unstated (Track~A), a reader turn taking \emph{agotado} literally asserts a fact the user omitted (a false positive, akin to hallucination). When the user states it (Track~B), the resulting note can look identical, so a reader discounting the form as routine grammar answers \emph{``not stated''} and loses that fact (a false negative, akin to forgetting). No reading rule alone resolves this precision--recall trade-off unless the note separates stated facts from grammatical defaults. Keeping the source breaks compaction's memory bound, while structured slots require fixed schemas, disrupt free-text workflows and invite hallucinated fills on unstated attributes \citep{usman2026phantomfill}.

We break this trade-off without training via a write--read convention and pre-eviction verification. Under the convention, the writer marks unstated attributes with paired forms, and a matching reader rule treats paired forms as unstated and single forms as verified facts. Before eviction, a likelihood check scores single forms against the message body and rewrites unsupported ones into paired forms. We make three contributions:
\begin{enumerate}
  \item \textbf{Problem and compaction audit.} We formalize the two coupled failures and quantify them on a new $300$-pair, five-language benchmark across seven compaction settings and two tasks.
  \item \textbf{Training-free repair.} Convention and verification together raise post-eviction F1 (which penalizes both errors) on all five models in both tasks; the convention drives nearly all of the gain, while verification safeguards the smallest writer.
  \item \textbf{Insights for memory design.} Bias arises only where grammar forces a choice (optional Japanese honorifics show none), and ablations explain why writer-only or reader-only fixes fall short.
\end{enumerate}

\section{Related Work}
\label{sec:related}
Our probes build on work on ambiguous gender and underspecification \citep{stanovsky2019evaluating,rarrick2023gate,gkovedarou2025gendel,savoldi2021gender,piergentili2023gente,wildenburg2024dust,menis2025assumed,doyen2026man,pranav2025glitter,satheesh2025ggbbq,staliunaite2026uncertainty}, which evaluates single-turn outputs with the source visible, whereas we study compacted memory \emph{after} eviction. Prompt and KV compression \citep{jiang2023llmlingua,zhang2023h2o,li2024snapkv,xiao2023streamingllm} and agent memory stores \citep{packer2023memgpt,chhikara2025mem0} assume notes preserve provenance \citep{buneman2001why}, while memory benchmarks \citep{maharana2024locomo,wu2025longmemeval} evaluate English histories with no forced choice. Still, LLMs favor self-generated text \citep{tan2024blinded,panickssery2024evaluators}, weigh dialogue roles unevenly \citep{pan2026user,laban2025lost}, misattribute memories \citep{ranjan2026reality} and fill unstated slots \citep{usman2026phantomfill}.

Unlike decoding-time context-aware or PMI methods \citep{shi2024cad,chen2026dynamicpmi,vanderpoel2022mutual,nandwani2023pmi}, sampling consistency \citep{farquhar2024semantic,quach2023conformal} or tool firewalls \citep{debenedetti2025camel,chennabasappa2025llamafirewall}, our verification leaves decoding unchanged and rewrites notes before eviction. Our check is closest to context attribution by ablation \citep{cohenwang2024contextcite}, but contrasts two inflected forms under a redacted source.

\section{Experimental Setup and Compaction Audit}
\label{sec:audit}

\subsection{Dataset and Scenarios}
We create all scenarios from scratch, with LLM assistance, so that each item forms an exact unstated--stated minimal pair. In each scenario, an English support or scheduling message $S_i$ with a ticket header (\texttt{[Booking \#ID]}) and a unisex sender name (\emph{Taylor Lin}) is translated at Turn~1 into the locale's working language and kept as the note $Z_i$. The main benchmark has $300$ pairs, $100$ per probe: Romance predicate gender ($25$ pairs each in Spanish, French, Italian and Portuguese), Mandarin sibling relative age (\emph{gege}/\emph{didi}) and Mandarin kinship lineage (\emph{waipo}/\emph{nainai}). Each \emph{unstated} input (\emph{``I was exhausted''}, \emph{``my grandmother''}) has a \emph{stated} twin adding an explicit cue: a pronoun tag (\emph{``Taylor Lin (she/her)''}), \emph{older}/\emph{younger}, or \emph{maternal}/\emph{paternal}. Messages are non-templated ($17$--$35$ words, six domains, $30$ unisex names) with no gendered, age or lineage words in any unstated input. Of the $180$ distinct messages, $120$ appear once and $60$ recur across the four Romance languages or across kinship nouns, yielding $300$ pairs.

\subsection{Compaction Settings, Tasks and Metrics}
\label{sec:settings}
\paragraph{Compaction settings.}
Our primary compacted setting is \emph{two-turn eviction}: in Turn~1 the assistant translates $S_i$ into the note $Z_i$; compaction evicts $S_i$ and keeps only $Z_i$ with the ticket ID; in Turn~2 the assistant answers a question or writes a handover from $Z_i$ alone. This isolates the loss of $S_i$ from distractor turns and retrieval errors, with all notes kept in the target language. Two references keeping $S_i$ (\emph{source kept, no note}; \emph{source and note both kept}) serve as upper bounds, while six additional settings add back multi-turn conversation, retrieval and user updates:
\begin{itemize}
  \item \textbf{Multi-turn compression (3):} \emph{six-turn sliding window} drops $S_i$ and appends three unrelated exchanges after $Z_i$; \emph{long chat} compacts a six-turn dialogue into one note; \emph{rolling summary} re-compacts $S_i$ and $Z_i$ into a two-sentence note.
  \item \textbf{Note retrieval (2):} \emph{BM25 top-4} queries all notes plus one closed earlier ticket per customer (a distractor lacking the tested attribute) by booking ID and sender name, returning four entries; \emph{ID-free retrieval} queries by sender name and email only.
  \item \textbf{Later update (1):} \emph{later user override} starts from $180$ unstated items and re-compacts after a follow-up that either adds no personal detail (the note should stay neutral) or first states the attribute opposite to any single form in the Turn-1 note.
\end{itemize}

\paragraph{Tasks.}
At Turn~2, the reader (the same model unless noted) runs two tasks: \textbf{question answering (QA)} asks it to pick form~A, form~B or \emph{not stated}, probing belief and abstention under a forced choice, while \textbf{agent handover (HO)} asks for a three-sentence English shift-handover email to test whether unstated assumptions leak into free text. QA uses exact option match; handovers are scored by matching form-specific words (\emph{she}, \emph{maternal grandmother}) against neutral wording (\emph{they}, \emph{grandmother}), validated by a cross-family LLM rater.\footnote{Agreement $98.8\%$, Cohen's $\kappa{=}0.97$.}

\paragraph{Models and AI tools.}
We evaluate \texttt{Qwen3.5-4B/9B/27B} \citep{qwen2026qwen35} and \texttt{Gemma-4-12B-it/31B-it} \citep{gemma2026gemma4} via vLLM \citep{kwon2023vllm} with greedy decoding (one output per input). Turn-1 writers and default readers share a system prompt against assuming unstated details; our reader rule replaces this default prompt. LLM assistants drafted the scenarios, and the experiment and analysis code was written with LLM coding assistance; the authors verified all designs, code and results and take full responsibility for the content.

\paragraph{Metrics.} In both tasks, the reader asserts form~A or form~B or abstains (\emph{not stated} in QA; neutral wording such as \emph{they} in handovers). An assertion is correct when the input stated that form. We report \textbf{precision} (\textbf{P}), the share of assertions that are correct; \textbf{recall} (\textbf{R}), the share of stated inputs recovered; and their harmonic mean \textbf{F1}. Because F1 gives no credit for abstaining on unstated inputs, we examine P alongside F1 and report \textbf{balanced accuracy} (\textbf{BA}), the average of R and the unstated abstention rate.

\begin{table}[t]
\centering
\caption{Compaction audit with default notes (five models pooled, \%). Rows 1--2 keep $S_i$; Rows 3--9 evict it. Row~5: $60$ pairs; Row~9: $180$ items.}
\label{tab:audit_regimes}
\footnotesize
\setlength{\tabcolsep}{0pt}
\renewcommand{\arraystretch}{0.87}
\begin{tabular*}{\columnwidth}{@{\extracolsep{\fill}}lcccccc@{}}
\toprule
& \multicolumn{3}{c}{\textbf{Turn-2 QA (\%)}} & \multicolumn{3}{c}{\textbf{Turn-2 Handover (\%)}} \\
\cmidrule(lr){2-4} \cmidrule(l){5-7}
\textbf{Compaction setting} & \textbf{P} & \textbf{R} & \textbf{F1} & \textbf{P} & \textbf{R} & \textbf{F1} \\
\midrule
1. Source kept, no note & 99.1 & 90.0 & 94.3 & 92.6 & 85.5 & 88.9 \\
2. Source and note both kept & 99.2 & 95.4 & 97.3 & 86.9 & 77.6 & 82.0 \\
\midrule
3. Two-turn eviction (main) & 89.8 & 34.2 & 49.5 & 62.1 & 23.6 & 34.2 \\
4. Six-turn sliding window & 96.9 & 29.6 & 45.4 & 63.6 & 12.1 & 20.4 \\
5. Long chat (6-turn transcript) & 80.0 & 1.3 & 2.6 & 34.0 & 5.7 & 9.7 \\
6. Rolling episodic summary & 70.6 & 13.1 & 22.1 & 42.5 & 12.3 & 19.1 \\
7. BM25 top-4 retrieval & 98.2 & 33.0 & 49.4 & 58.7 & 19.9 & 29.8 \\
8. ID-free retrieval & 98.3 & 26.8 & 42.1 & 63.2 & 21.7 & 32.3 \\
\midrule
9. Later user override & 94.7 & 60.1 & 73.6 & 80.5 & 47.2 & 59.5 \\
\bottomrule
\end{tabular*}
\end{table}

\subsection{Compaction Audit Results}
\paragraph{Stated-fact loss across settings (Table~\ref{tab:audit_regimes}).}
Keeping $S_i$ (Rows~1--2) yields high recall and near-ceiling precision on both tasks. Evicting $S_i$ (Rows~3--9) splits the failure mode by task. In QA, where a \emph{not stated} option exists, cautious readers treat inflected forms in $Z_i$ as routine grammar and abstain: precision stays high ($89.8\%$), yet recall drops by nearly two-thirds (from $90.0\%$ to $34.2\%$, Row~3). In free-text handovers, where no abstain option exists, readers lose both recall ($23.6\%$) via neutral phrasing and precision ($62.1\%$) as unsupported inflections leak into gendered pronouns (\emph{she}/\emph{he}). Deeper compression compounds this loss because summaries drop surviving cues (Rows~5--6), whereas retrieval distractors add little damage beyond eviction (Rows~7--8).

\paragraph{Bias on unstated inputs.}
Default notes are biased wherever target grammar \emph{forces} a personal attribute. On unstated inputs, writers pick a single form in $52.5\%$ of notes ($45$--$59\%$ per model) and both forms in $4.9\%$: single-form guessing is $74\%$ for gender and $78\%$ for sibling age, but only $5\%$ for lineage. Lineage reveals a second provenance failure: writers avoid guessing by using generic \emph{zumu}/\emph{zufu}/\emph{shushu}, yet also write those same words for $132$ of $180$ stated paternal relatives. This hides a stated fact behind an ambiguous term, and default readers recover it from only $7$ of the $132$ notes. On a separate $40$-item probe, evicting the source raises QA probability for the written form by about $21$ percentage points (pp) on average, and scaling from $4$B to $31$B does not reduce this shift.

By contrast, no bias appears where grammar is optional: on $80$ Japanese items where honorifics may, but need not, mark seniority, writers encode it in $14.5\%$ of notes and post-eviction shifts stay below $3$~pp on every model. Vulnerable language pairs are thus predictable from grammar: bias arises when the target language forces a distinction that the source leaves open.

\section{Method and Repair Evaluation}
\label{sec:methods}
\begin{table*}[t]
\centering
\caption{Two-turn eviction results per model (\%). Full method = Methods~1+2; w/o verification = Method~1; BA = balanced accuracy. Bold: best F1 per column; $\dagger$: above the default note at $p{<}0.05$ (paired bootstrap).}
\label{tab:permodel}
\footnotesize
\setlength{\tabcolsep}{0pt}
\renewcommand{\arraystretch}{0.87}
\begin{tabular*}{\textwidth}{@{\extracolsep{\fill}}lcccccccccccccccccccc@{\hspace{3pt}}}
\toprule
& \multicolumn{4}{c}{\textbf{Qwen3.5-4B}} & \multicolumn{4}{c}{\textbf{Qwen3.5-9B}} & \multicolumn{4}{c}{\textbf{Qwen3.5-27B}} & \multicolumn{4}{c}{\textbf{Gemma-4-12B}} & \multicolumn{4}{c}{\textbf{Gemma-4-31B}} \\
\cmidrule(lr){2-5} \cmidrule(lr){6-9} \cmidrule(lr){10-13} \cmidrule(lr){14-17} \cmidrule(l){18-21}
\textbf{Configuration} & P & R & F1 & BA & P & R & F1 & BA & P & R & F1 & BA & P & R & F1 & BA & P & R & F1 & BA \\
\midrule
\multicolumn{21}{@{}l}{\textbf{Question answering (QA)}} \\
1. Default target-language note & 89.2 & 19.3 & 31.8 & 58.5 & 97.4 & 37.3 & 54.0 & 68.3 & 93.4 & 33.0 & 48.8 & 65.5 & 81.2 & 57.7 & 67.4 & 72.2 & 98.6 & 23.7 & 38.2 & 61.7 \\
2. Reader rule only (default notes) & 55.7 & 47.3 & 51.2\rlap{$^{\dagger}$} & 55.7 & 93.7 & 39.7 & 55.7 & 68.7 & 82.2 & 40.0 & 53.8\rlap{$^{\dagger}$} & 67.2 & 79.6 & 66.3 & 72.4\rlap{$^{\dagger}$} & 74.8\rlap{$^{\dagger}$} & 73.4 & 80.0 & 76.6\rlap{$^{\dagger}$} & 75.8\rlap{$^{\dagger}$} \\
3. Full method (ours) & 65.7 & 53.7 & \textbf{59.1}\rlap{$^{\dagger}$} & 63.3\rlap{$^{\dagger}$} & 98.8 & 52.7 & \textbf{68.7}\rlap{$^{\dagger}$} & 76.0\rlap{$^{\dagger}$} & 96.2 & 59.7 & \textbf{73.7}\rlap{$^{\dagger}$} & 79.7\rlap{$^{\dagger}$} & 95.1 & 64.3 & \textbf{76.7}\rlap{$^{\dagger}$} & 81.0\rlap{$^{\dagger}$} & 99.6 & 84.7 & \textbf{91.5}\rlap{$^{\dagger}$} & 92.3\rlap{$^{\dagger}$} \\
4. \ \ w/o verification & 58.8 & 53.7 & 56.1\rlap{$^{\dagger}$} & 58.5 & 98.1 & 52.7 & 68.5\rlap{$^{\dagger}$} & 75.8\rlap{$^{\dagger}$} & 96.2 & 59.7 & \textbf{73.7}\rlap{$^{\dagger}$} & 79.7\rlap{$^{\dagger}$} & 94.6 & 64.3 & 76.6\rlap{$^{\dagger}$} & 80.8\rlap{$^{\dagger}$} & 99.6 & 84.7 & \textbf{91.5}\rlap{$^{\dagger}$} & 92.3\rlap{$^{\dagger}$} \\
\midrule
\multicolumn{21}{@{}l}{\textbf{Agent handover (HO)}} \\
1. Default target-language note & 46.8 & 24.3 & 32.0 & 50.0 & 61.0 & 36.0 & 45.3 & 56.7 & 70.4 & 25.3 & 37.3 & 57.3 & 74.2 & 23.0 & 35.1 & 57.5 & 77.8 & 9.3 & 16.7 & 53.3 \\
2. Reader rule only (default notes) & 53.4 & 34.3 & 41.8\rlap{$^{\dagger}$} & 53.0\rlap{$^{\dagger}$} & 62.6 & 43.0 & 51.0\rlap{$^{\dagger}$} & 58.8 & 82.7 & 38.3 & 52.4\rlap{$^{\dagger}$} & 65.2\rlap{$^{\dagger}$} & 88.7 & 34.0 & 49.2\rlap{$^{\dagger}$} & 64.8\rlap{$^{\dagger}$} & 95.0 & 32.0 & 47.9\rlap{$^{\dagger}$} & 65.2\rlap{$^{\dagger}$} \\
3. Full method (ours) & 61.5 & 40.0 & \textbf{48.5}\rlap{$^{\dagger}$} & 58.8\rlap{$^{\dagger}$} & 78.2 & 45.3 & \textbf{57.4}\rlap{$^{\dagger}$} & 66.3\rlap{$^{\dagger}$} & 96.1 & 49.0 & \textbf{64.9}\rlap{$^{\dagger}$} & 73.5\rlap{$^{\dagger}$} & 95.8 & 38.0 & \textbf{54.4}\rlap{$^{\dagger}$} & 68.5\rlap{$^{\dagger}$} & 100.0 & 40.0 & 57.1\rlap{$^{\dagger}$} & 70.0\rlap{$^{\dagger}$} \\
4. \ \ w/o verification & 58.3 & 40.0 & 47.4\rlap{$^{\dagger}$} & 57.0\rlap{$^{\dagger}$} & 75.8 & 46.0 & 57.3\rlap{$^{\dagger}$} & 65.7\rlap{$^{\dagger}$} & 96.1 & 49.0 & \textbf{64.9}\rlap{$^{\dagger}$} & 73.5\rlap{$^{\dagger}$} & 95.8 & 37.7 & 54.1\rlap{$^{\dagger}$} & 68.3\rlap{$^{\dagger}$} & 100.0 & 40.3 & \textbf{57.5}\rlap{$^{\dagger}$} & 70.2\rlap{$^{\dagger}$} \\
\midrule
\multicolumn{21}{@{}l}{\textbf{Unstated inputs (\%, $\downarrow$): default $\to$ w/o verification $\to$ full method}} \\
Note carries a single form & \multicolumn{4}{c}{58.3 $\to$ 53.7 $\to$ \textbf{20.7}} & \multicolumn{4}{c}{51.3 $\to$ 23.7 $\to$ \textbf{10.3}} & \multicolumn{4}{c}{58.7 $\to$ 4.3 $\to$ \textbf{3.7}} & \multicolumn{4}{c}{49.0 $\to$ 11.0 $\to$ \textbf{9.0}} & \multicolumn{4}{c}{45.3 $\to$ 8.0 $\to$ \textbf{5.7}} \\
Reader asserts a form, QA & \multicolumn{4}{c}{2.3 $\to$ 36.7 $\to$ 27.0} & \multicolumn{4}{c}{0.7 $\to$ 1.0 $\to$ 0.7} & \multicolumn{4}{c}{2.0 $\to$ 0.3 $\to$ 0.3} & \multicolumn{4}{c}{13.3 $\to$ 2.7 $\to$ 2.3} & \multicolumn{4}{c}{0.3 $\to$ 0.0 $\to$ 0.0} \\
Reader asserts a form, HO & \multicolumn{4}{c}{24.3 $\to$ 26.0 $\to$ 22.3} & \multicolumn{4}{c}{22.7 $\to$ 14.7 $\to$ 12.7} & \multicolumn{4}{c}{10.7 $\to$ 2.0 $\to$ 2.0} & \multicolumn{4}{c}{8.0 $\to$ 1.0 $\to$ 1.0} & \multicolumn{4}{c}{2.7 $\to$ 0.0 $\to$ 0.0} \\
\bottomrule
\end{tabular*}
\end{table*}

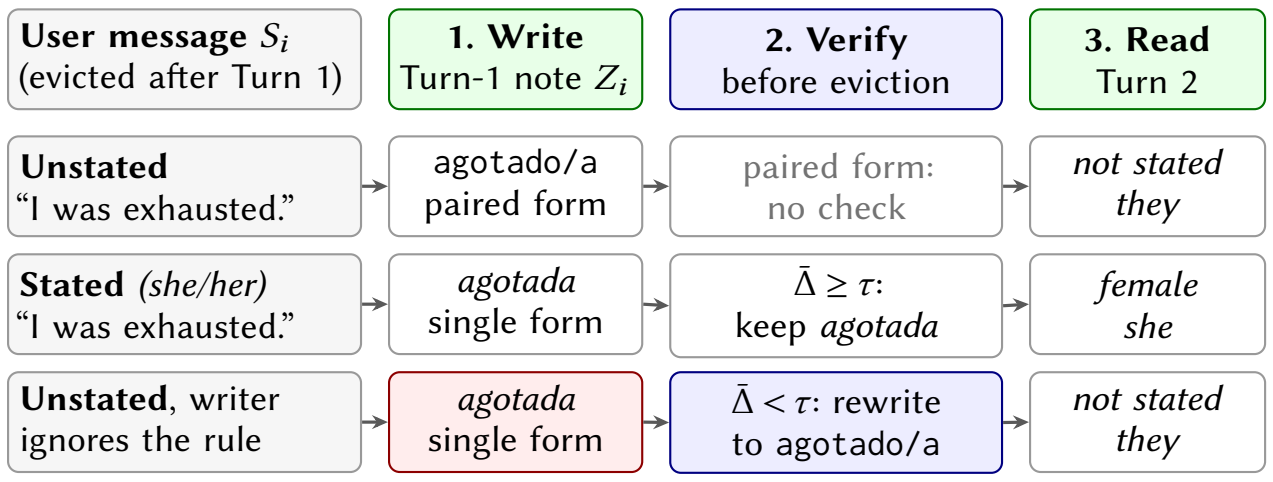
\begin{figure}[t]
\centering
\begin{tikzpicture}[
  font=\fontsize{6.9}{7.8}\selectfont\sffamily,
  cell/.style={draw=black!40, rounded corners=2pt, inner sep=2.2pt, align=center, minimum height=0.66cm, anchor=north west},
  cin/.style={cell, align=left, fill=gray!7, text width=2.18cm},
  cw/.style={cell, text width=1.52cm},
  cv/.style={cell, text width=2.04cm},
  cr/.style={cell, text width=1.40cm},
  m1/.style={fill=green!9, draw=green!45!black},
  m2/.style={fill=blue!7, draw=blue!50!black},
  bad/.style={fill=red!7, draw=red!55!black},
  arr/.style={-{Stealth[length=3.0pt,width=2.6pt]}, semithick, draw=black!65}
]
\node[cin] (h0) at (0,0) {\textbf{User message $S_i$}\\(evicted after Turn 1)};
\node[cw, m1] (h1) at (2.52cm,0) {\textbf{1. Write}\\Turn-1 note $Z_i$};
\node[cv, m2] (h2) at (4.38cm,0) {\textbf{2. Verify}\\before eviction};
\node[cr, m1] (h3) at (6.76cm,0) {\textbf{3. Read}\\Turn 2};
\node[cin] (a0) at (0,-0.84cm) {\textbf{Unstated}\\``I was exhausted.''};
\node[cw] (a1) at (2.52cm,-0.84cm) {\texttt{agotado/a}\\paired form};
\node[cv, text=black!55] (a2) at (4.38cm,-0.84cm) {paired form:\\no check};
\node[cr] (a3) at (6.76cm,-0.84cm) {\emph{not stated}\\\emph{they}};
\node[cin] (b0) at (0,-1.62cm) {\textbf{Stated} \emph{(she/her)}\\``I was exhausted.''};
\node[cw] (b1) at (2.52cm,-1.62cm) {\emph{agotada}\\single form};
\node[cv] (b2) at (4.38cm,-1.62cm) {$\bar{\Delta}\,{\geq}\,\tau$:\\keep \emph{agotada}};
\node[cr] (b3) at (6.76cm,-1.62cm) {\emph{female}\\\emph{she}};
\node[cin] (c0) at (0,-2.40cm) {\textbf{Unstated}, writer\\ignores the rule};
\node[cw, bad] (c1) at (2.52cm,-2.40cm) {\emph{agotada}\\single form};
\node[cv, m2] (c2) at (4.38cm,-2.40cm) {$\bar{\Delta}\,{<}\,\tau$: rewrite\\to \texttt{agotado/a}};
\node[cr] (c3) at (6.76cm,-2.40cm) {\emph{not stated}\\\emph{they}};
\foreach \r in {a,b,c}{
  \draw[arr] (\r0.east) -- (\r1.west);
  \draw[arr] (\r1.east) -- (\r2.west);
  \draw[arr] (\r2.east) -- (\r3.west);
}
\end{tikzpicture}
\Description{A grid with three steps as columns (1. Write the Turn-1 note, 2. Verify before eviction, 3. Read at Turn 2) and three cases as rows. Unstated input: the writer uses the paired form agotado/a, verification skips it, and the reader answers not stated or writes they. Stated input (she/her): the writer uses the single form agotada, verification keeps it, and the reader answers female or writes she. Unstated input where the writer ignores the rule and writes agotada: verification rewrites it to agotado/a, and the reader answers not stated or writes they.}
\caption{Our repair on three cases: the write--read \textbf{convention} (Method~1, green) and the pre-eviction \textbf{verification} (Method~2, purple).}
\label{fig:method}
\end{figure}

Fig.~\ref{fig:method} illustrates our repair: a write--read \emph{convention} (Method~1) paired with compaction-time \emph{verification} (Method~2).

\subsection{Method 1: Coupled Write--Read Convention}
\begin{enumerate}
  \item \textbf{Turn-1 writer rule.} Before multilingual compaction, an instruction appended to the translation request tells the writer to check whether $S_i$ states the attribute for each person via a pronoun tag such as \emph{(she/her)} or kinship modifiers (\emph{older}/\emph{younger}, \emph{maternal}/\emph{paternal}). Six word-level examples illustrate the rule: stated attributes map to a single form (\emph{(he/him)}$\to$\texttt{agotado}, \emph{older brother}$\to$\emph{gege}), whereas unstated ones map to a paired form (\texttt{agotado/a}, \emph{gege/didi}, \emph{waipo/nainai}).
  \item \textbf{Turn-2 reader rule.} After $S_i$ is evicted, a short system prompt and a one-line header above $Z_i$ specify how to read $Z_i$: paired forms mean the user left the attribute open, so the reader answers \emph{not stated} in QA and uses neutral wording (singular \emph{they} or a neutral kinship term) in handovers, whereas single forms mark verified facts.
\end{enumerate}

\subsection{Method 2: Compaction Likelihood Check}
Small writers (e.g., \texttt{Qwen3.5-4B}) or third-party writers may ignore the Turn-1 rule and translate an unstated \emph{``I was exhausted''} as \emph{``Taylor Lin estaba agotada''}, which the reader rule would then trust. Verification (Method~2) catches such slips with a likelihood check right before $S_i$ is evicted. Comparing the likelihoods of $Z_i$ and opposite-form $Z_{i,\text{opp}}$ (\emph{agotada}$\to$\emph{agotado}) directly under $S_i$ would confound message evidence with priors over words or sender names; the check isolates the body's gain:
\begin{enumerate}
  \item \textbf{Form swap.} Notes already using paired forms or carrying an explicit non-obligatory qualifier (a pronoun tag or explicit lineage/age modifier) pass unchanged. If a note $Z_i$ lacks such a qualifier and contains $k \ge 1$ single-form spans from a target-language form-pair list, we swap all $k$ spans to their opposite form for $Z_{i,\text{opp}}$ (partial swaps break grammatical agreement).
  \item \textbf{Body redaction.} $S_{\emptyset}$ replaces the body of $S_i$ with a placeholder while retaining the ticket ID and sender name.
  \item \textbf{Log-likelihood ratio.} Four forward passes of the Turn-1 model, without token generation, score the four (message, note) pairs to yield the per-span log-likelihood ratio
  \begin{equation}
  \bar{\Delta}(Z_i) = \frac{1}{k}\left[\ln \frac{P(Z_i \mid S_i)}{P(Z_i \mid S_{\emptyset})} - \ln \frac{P(Z_{i,\text{opp}} \mid S_i)}{P(Z_{i,\text{opp}} \mid S_{\emptyset})}\right]
  \label{eq:cppmi}
  \end{equation}
  In Eq.~\ref{eq:cppmi}, shared words, name priors and base form frequencies cancel out: when $S_i$ says only \emph{``I was exhausted''}, restoring the body ideally shifts \emph{agotada} and \emph{agotado} similarly ($\bar{\Delta}$ small), though some unstated notes still pass $\tau$ or lack a matched span (Table~\ref{tab:permodel}); when $S_i$ specifies \emph{``Taylor Lin (she/her) was exhausted''}, restoring the body favors \emph{agotada} ($\bar{\Delta}\gg 0$).
  \item \textbf{Rewrite.} If $\bar{\Delta}(Z_i) < \tau$, string replacement converts all $k$ spans into paired forms (\emph{agotada}$\to$\texttt{agotado/a}) before evicting $S_i$. With $\tau{=}1.0$ (fixed a priori, untuned; Section~\ref{sec:eval} varies it), a single form survives only when the message body raises the relative probability of the written form by at least $e\approx 2.7\times$ per span.
\end{enumerate}

\subsection{Repair Evaluation}
\label{sec:eval}
\noindent \textbf{Main comparison (Table~\ref{tab:permodel}).}
The full method (Row~3) improves F1 and balanced accuracy over default notes (Row~1) on every model and task. On default notes, discounting (low recall) is the dominant failure and grows above $9$B: \texttt{Gemma-4-31B} achieves high precision by treating most target-language forms as routine grammar, keeping its handover recall below $10\%$ and its default F1 below that of \texttt{Gemma-4-12B} in Tables~\ref{tab:permodel} and~\ref{tab:settings}. Applying the reader rule alone (Row~2) helps cautious readers recover facts, yet lowers precision in QA because it trusts unsupported guesses; the full method outperforms it on every model and task.

The convention works to the extent that the writer follows it. The four $9$B--$31$B models recover recall at high precision under Method~1 alone (Row~4). By contrast, \texttt{Qwen3.5-4B} still leaves $53.7\%$ of notes on unstated inputs single-form (default: $58.3\%$), which the reader rule trusts. Verification cuts this rate from $53.7\%$ to $20.7\%$ (the remaining single-form notes either pass $\tau{=}1.0$ or lack a matched span) and lifts the full method above the convention alone on \texttt{Qwen3.5-4B} in both tasks (QA precision $+6.9$ points), whereas on $9$B--$31$B it changes F1 by at most one point. Even with verification, \texttt{Qwen3.5-4B} asserts a form on $27.0\%$ of unstated QA inputs, up from $2.3\%$ with default notes, the only one of the ten model--task columns in Table~\ref{tab:permodel} where assertions rise.

The risk lies in the reader rule. Under the default reader, our notes (Table~\ref{tab:ablation}, ``w/o Turn-2 reader rule'') still beat default notes (Table~\ref{tab:permodel}, Row~1) in QA-F1 on every model, and \texttt{Qwen3.5-4B} asserts a form on only $2.0\%$ of unstated QA inputs. We thus recommend enabling the reader rule only for writers that rarely leave unstated inputs single-form after verification (at most $10.3\%$ on $9$B--$31$B, vs.\ $20.7\%$ on $4$B).

\begin{table}[t]
\centering
\caption{QA-F1 ablations (\%) under two-turn eviction. Bold: best non-indented row.}
\label{tab:ablation}
\footnotesize
\setlength{\tabcolsep}{0pt}
\renewcommand{\arraystretch}{0.87}
\begin{tabular*}{\columnwidth}{@{\extracolsep{\fill}}lccccc@{}}
\toprule
& \multicolumn{3}{c}{\textbf{Qwen3.5}} & \multicolumn{2}{c}{\textbf{Gemma-4}} \\
\cmidrule(lr){2-4} \cmidrule(l){5-6}
\textbf{Configuration} & \textbf{4B} & \textbf{9B} & \textbf{27B} & \textbf{12B} & \textbf{31B} \\
\midrule
\multicolumn{6}{@{}l}{\textbf{QA-F1$\uparrow$}} \\
Convention only (w/o verification) & 56.1 & 68.5 & \textbf{73.7} & 76.6 & \textbf{91.5} \\
\quad w/o Turn-1 examples & 47.2 & 41.9 & 46.4 & 76.1 & 81.3 \\
Full method (Methods~1+2) & \textbf{59.1} & \textbf{68.7} & \textbf{73.7} & \textbf{76.7} & \textbf{91.5} \\
\quad w/o Turn-2 reader rule & 45.1 & 66.7 & 69.8 & 76.0 & 53.7 \\
\midrule
One-sentence writer, default reader & 22.2 & 50.7 & 61.1 & 66.7 & 25.6 \\
\midrule
Reader rule only, default notes & 51.2 & 55.7 & 53.8 & 72.4 & 76.6 \\
Verification + reader, default notes & \textbf{52.1} & \textbf{56.0} & \textbf{53.9} & \textbf{73.1} & \textbf{79.2} \\
\bottomrule
\end{tabular*}
\end{table}

\noindent \textbf{Ablations (Table~\ref{tab:ablation}).}
Both parts of the convention are essential: removing either lowers QA-F1 on every model. Without the six Turn-1 writer examples, models over-apply paired forms to stated attributes (no single form on $47.1\%$ of stated inputs, vs.\ $25.3\%$ with examples); Turn-2 readers then abstain, cutting F1 most sharply on \texttt{Qwen3.5-9B} and \texttt{27B}. Without the Turn-2 reader rule, default readers revert to discounting single forms as grammar, hurting \texttt{Gemma-4-31B} and \texttt{Qwen3.5-4B} most.

A simpler one-sentence instruction also falls short. Instructed only to keep unstated attributes ambiguous with paired forms (four examples, with no rule for stated inputs), writers again pair stated attributes: under the default reader (Table~\ref{tab:ablation}), QA-F1 falls below default notes (Table~\ref{tab:permodel}, Row~1) on four of five models and below our notes under the same reader on all five.

\noindent \textbf{Verification without the writer rule.}
Verification alone cannot make an unchanged writer compliant, though it still applies whenever a writer cannot be modified (e.g., a third party) and $S_i$ remains available at eviction. Run before the reader rule on default notes, it improves QA-F1 on all five models (Table~\ref{tab:ablation}, last two rows) but stays below the full method on each.

\noindent \textbf{Generalization across compaction settings (Table~\ref{tab:settings}).}
Our repair generalizes across the six multi-turn, retrieval and update settings: the full method raises QA-F1 in $28$ of $30$ model--setting cells (Table~\ref{tab:settings}) and handover F1 in $28$ of $30$. All four exceptions appear in \emph{later user override}, where a fresh follow-up states the attribute right before re-compaction, allowing mid-scale default notes to score well.

\begin{table}[t]
\centering
\caption{QA-F1 (\%), default notes (\textbf{Def.}) vs.\ \textbf{full method} (\textbf{Full}), six other settings. Bold: higher of the pair; $\dagger$: $p{<}0.05$.}
\label{tab:settings}
\footnotesize
\setlength{\tabcolsep}{0pt}
\renewcommand{\arraystretch}{0.87}
\begin{tabular*}{\columnwidth}{@{\extracolsep{\fill}}lcccccccccc@{}}
\toprule
& \multicolumn{2}{c}{\textbf{Q-4B}} & \multicolumn{2}{c}{\textbf{Q-9B}} & \multicolumn{2}{c}{\textbf{Q-27B}} & \multicolumn{2}{c}{\textbf{G-12B}} & \multicolumn{2}{c}{\textbf{G-31B}} \\
\cmidrule(lr){2-3} \cmidrule(lr){4-5} \cmidrule(lr){6-7} \cmidrule(lr){8-9} \cmidrule(l){10-11}
\textbf{Setting} & Def. & Full\hphantom{$^{\dagger}$} & Def. & Full\hphantom{$^{\dagger}$} & Def. & Full\hphantom{$^{\dagger}$} & Def. & Full\hphantom{$^{\dagger}$} & Def. & Full\hphantom{$^{\dagger}$} \\
\midrule
Six-turn sliding window & 27.1 & \textbf{47.2}$^{\dagger}$ & 51.4 & \textbf{58.5}$^{\dagger}$ & 40.5 & \textbf{64.6}$^{\dagger}$ & 63.4 & \textbf{78.1}$^{\dagger}$ & 38.7 & \textbf{88.5}$^{\dagger}$ \\
Long chat (6-turn transcript) & 3.3 & \textbf{34.2}$^{\dagger}$ & 0.0 & \textbf{64.4}$^{\dagger}$ & 0.0 & \textbf{65.2}$^{\dagger}$ & 9.4 & \textbf{76.0}$^{\dagger}$ & 0.0 & \textbf{90.3}$^{\dagger}$ \\
Rolling episodic summary & 16.0 & \textbf{41.9}$^{\dagger}$ & 38.2 & \textbf{52.9}$^{\dagger}$ & 15.1 & \textbf{55.2}$^{\dagger}$ & 33.5 & \textbf{68.9}$^{\dagger}$ & 0.7 & \textbf{86.5}$^{\dagger}$ \\
BM25 top-4 retrieval & 43.8 & \textbf{54.5}$^{\dagger}$ & 51.4 & \textbf{69.7}$^{\dagger}$ & 50.1 & \textbf{83.7}$^{\dagger}$ & 58.7 & \textbf{75.6}$^{\dagger}$ & 41.7 & \textbf{80.4}$^{\dagger}$ \\
ID-free retrieval & 27.1 & \textbf{49.3}$^{\dagger}$ & 42.0 & \textbf{66.8}$^{\dagger}$ & 43.8 & \textbf{78.7}$^{\dagger}$ & 55.3 & \textbf{69.5}$^{\dagger}$ & 39.6 & \textbf{78.5}$^{\dagger}$ \\
Later user override & 66.4 & \textbf{74.1}$^{\dagger}$ & 79.3 & \textbf{80.7}\hphantom{$^{\dagger}$} & \kern-1pt\textbf{94.8}\rlap{$^{\dagger}$}\kern1pt & \kern1pt 87.1\hphantom{$^{\dagger}$}\kern-1pt & \textbf{80.8} & 77.7\hphantom{$^{\dagger}$} & 27.8 & \textbf{76.9}$^{\dagger}$ \\
\bottomrule
\end{tabular*}
\end{table}

Gains are largest under deep compression (\emph{long chat}, \emph{rolling summary}). Default \emph{long-chat} summaries mostly drop the stated attribute, driving their QA-F1 near zero, whereas the full method restates the convention at each re-compaction.

\noindent \textbf{Robustness and latency.}
Gains hold without tuning $\tau$ or matching the scorer to the writer. Using a stricter $\tau{=}2.5$, calibrating $\tau$ via leave-one-language-out selection, or scoring notes from all five writers with \texttt{Qwen3.5-4B} each beat default notes on every model and task. Compliance also holds under default stochastic decoding: across $16$ rollouts per input, the share of unstated inputs avoiding a single form stays within $5.7$~pp of the greedy rate on every model.

The convention transfers beyond the six writer examples (Spanish, French and Mandarin). On the $100$ gender pairs, Italian and Portuguese gain as much in handovers as Spanish and French (Method~1 F1 $+21.8$ vs.\ $+18.4$); in QA, where default F1 already reaches $86$--$87$, both gain little ($+1.7$ vs.\ $+3.6$). On a held-out category (Mandarin older vs.\ younger paternal uncle, $25$ adapted pairs), Method~1 raises pooled QA-F1 from $32.9$ to $61.3$ and handover F1 from $14.8$ to $51.2$, with four of five models gaining per task.

Runtime overhead is small. On one H100 GPU (vLLM, batch size $50$), the Turn-1 convention raises note generation from $37$ to $50$~ms per note with no added Turn-2 latency. Verification takes $91$~ms per checked note (timed on $50$ unstated Romance inputs per model), runs on the $41.3\%$ of convention notes lacking paired forms or explicit qualifiers, and executes asynchronously at eviction.

\noindent \textbf{Limitations.}
Our synthetic benchmark uses mainly explicit attribute cues. Writer examples and form lists cover only three categories, and we test two open-weight model families up to 31B.

\section{Conclusion}
\label{sec:conclusion}
When an assistant compacts a user message into a target-language note and evicts the original, the note obscures whether a grammar-forced form reflects user input or grammar, so later turns assert unstated attributes or, far more often, discount stated ones. Our training-free write--read convention marks unstated forms in the note, pairs with a pre-eviction check of single forms against the message body, and raises post-eviction F1 by $9$--$53$ points in QA and $12$--$40$ points in handovers across all five models. Compacted notes should therefore record whether each grammar-forced form was user-stated. Future work should extend this design to implied attributes, additional languages and larger or proprietary models.

\section*{Ethical Considerations}
This study audits how obligatory grammatical distinctions in machine-generated summaries, such as gender or kinship lineage, propagate after context compaction into downstream question answering and agent handovers. All benchmark scenarios and stated-attribute pairs are synthetically authored from fictional support and scheduling situations with balanced unisex given names; no real customer records or human-subject annotations were collected. Reducing unsupported attribute assertions while retaining user-stated pronouns and family relations helps prevent unverified demographic assumptions in compacted notes. Throughout our evaluation, singular \emph{they} and \emph{not stated} answers, like paired forms, are scored as accurate outputs.

The reader rule should be enabled only for writers that follow the convention (Section~\ref{sec:eval}). Paired forms such as \texttt{agotado/a} are binary: our benchmark states only \emph{she/her} and \emph{he/him}, and how to record a stated non-binary identity remains open.

\bibliographystyle{ACM-Reference-Format}
\bibliography{references}

\end{document}